\def\year{2022}\relax
\documentclass[letterpaper]{article} 
\usepackage{aaai22}  
\usepackage{times}  
\usepackage{helvet}  
\usepackage{courier}  
\usepackage[hyphens]{url}  
\usepackage{graphicx} 
\usepackage{natbib}  
\usepackage{caption} 
\DeclareCaptionStyle{ruled}{labelfont=normalfont,labelsep=colon,strut=off} 
\usepackage{soul}
\usepackage{nameref}
\usepackage{amssymb,amsmath}
\usepackage{comment,enumerate}
\usepackage{graphicx}
\usepackage{tablefootnote}
\usepackage{color,url}
\usepackage{multirow}
\usepackage{bbm}
\usepackage{subcaption}
\usepackage{algorithm}
\usepackage{algpseudocode}
\usepackage{makecell}
\usepackage{stfloats}
\usepackage{amsthm,amssymb,amsmath}

\usepackage[T1]{fontenc}

\begin{document}

\title{Everything is Varied: The Surprising Impact of Individual Variation on ML Robustness in Medicine}
\author{Andrea Campagner$^1$, Lorenzo Famiglini$^1$, Anna Carobene$^3$, Federico Cabitza$^{1,2}$ }
\affiliations{$^1$ Dipartimento di Informatica, Sistemistica e Comunicazione, University of Milano-Bicocca, Milano, Italy \\
$^2$ IRCCS Istituto Ortopedico Galeazzi, Milano, Italy \\
$^3$ IRCCS Ospedale San Raffaele, Milano, Italy
}

\maketitle

\begin{abstract}
In medical settings, Individual Variation (IV) refers to variation that is due not to population differences or errors, but rather to within-subject variation, that is the intrinsic and characteristic patterns of variation pertaining to a given instance or the measurement process. While taking into account IV has been deemed critical for proper analysis of medical data, this source of uncertainty and its impact on robustness have so far been neglected in Machine Learning (ML). To fill this gap, we look at how IV affects ML performance and generalization and how its impact can be mitigated. Specifically, we provide a methodological contribution to formalize the problem of IV in the statistical learning framework and, through an experiment based on one of the largest real-world laboratory medicine datasets for the problem of COVID-19 diagnosis, we show that: 1) common state-of-the-art ML models are severely impacted by the presence of IV in data; and 2) advanced learning strategies, based on data augmentation and data imprecisiation, and proper study designs can be effective at improving robustness to IV. Our findings demonstrate the critical relevance of correctly accounting for IV to enable safe deployment of ML in clinical settings.
\end{abstract}

\section{Introduction}
\label{sec:intro}
In recent years, the interest toward the application of Machine Learning (ML) methods and systems to the development of decision support systems in clinical settings has been steadily increasing~\cite{benjamens2020state}. This interest has been mainly driven by the promising results obtained and reported by these systems in academic research for different tasks~\cite{aggarwal2021diagnostic,yin2021role}

Despite these promising results, the adoption of ML-based systems in real-world clinical settings has been lagging behind~\cite{wilkinson2020time}, 
with these systems often failing to meet the expectations and requirements needed for safe deployment in clinical settings~\cite{andaur2021risk,futoma2020myth}, a concept that has been termed the \emph{last mile of implementation}~\cite{coiera2019last}
. While reasons behind the gaps in this ``last mile'' are numerous, among them we recall the inability of ML systems to reliably generalize in new contexts and settings~\cite{beam2020challenges,christodoulou2019systematic}, as well as their lack of robustness and susceptibility to variation in data, leading to poorer performance in real settings~\cite{li2019diagnosis} and, ultimately, to what has been called the \emph{replication crisis} of ML in medicine~\cite{coiera2018does}.

In the ML literature, the notion of variation has usually been associated with variance in the population data distribution (that is, as it relates to either the larger reference population, or a sample taken from this latter), due to the presence of outliers or anomalies~\cite{akoglu2021anomaly}, out-of-distribution instances~\cite{adila2022understanding,morteza2022provable} or concept/co-variate shifts and drifts~\cite{liu2021stable,rabanser2019failing}. While these forms of variation are certainly relevant, however they are not the only ones that can arise in real-world settings: indeed, another source of variation
in data is the so-called \emph{individual variation} (IV)~\cite{fraser2001biological}, which is especially common in laboratory data or other physiological signals and biomarkers, and more generally in every phenomenon whose manifestations can exhibit time-varying patterns.

IV denotes between-subject variation that is not due to population differences or errors, but rather to the intrinsic and characteristic (that is, individual) patterns of variation pertaining to single instances, that is to within-subject variation~\cite{plebani2015biological}; and more specifically it relates to two possible sources of variation: either the feature values for a given subject or patient, what is called \emph{biological variation} (BV)~\cite{plebani2015biological}; or the measurement process and instrument itself, i.e., what is called \emph{analytical variation} (AV). The presence of IV entails~\cite{badrick2021biological} that for each individual one can identify a ``subject average'' or central tendency (\emph{homeostatic point}~\cite{fraser2001biological}) arising from such factors as personal characteristic of the individuals themselves (e.g., genetic characteristics, age, phenotypic elements such as diet and physical activity) or of the measurement instrument (e.g., instrument calibration), as well as a distribution of possible values, whose uncertainty is represented by the extent of the IV: crucially, only a snapshot (i.e., a sample) from this distribution can be accessed at any moment.

While the potential impact of IV on computer-supported diagnosis has been known for a while (for instance, in~\cite{spodick1997computer} authors reported that ``computer interpretations of electrocardiograms recorded 1 minute apart were significantly (grossly) different in 4 of 10 cases''), only conjectures have so far been produced to estimate its extent. Nonetheless, IV has two strong implications for ML applications. First, ML models trained on data affected by IV, even highly accurate ones, can fail to be robust and properly generalize not only to new patients, but also to the same patients observed in slightly different conditions: for example, an healthy patient could indeed be classified as healthy with respect to the features actually observed for them, while they could have been classified as non-healthy for a slightly different set of feature values, which nevertheless would still be totally compatible with the distribution due to IV\footnote{As we show in the following, this setting is a generalization of the usual one adopted in ML theory~\cite{shalev2014understanding}: not only we assume that the best model could have less than perfect accuracy, but we also assume that any instance is represented as a distribution of vectors possibly lying in opposite sides of the decision boundary.}.
Second, differently from distribution-related variation, collecting additional data samples, which has been considered a primary factor in the continued improvement of ML systems, can help only marginally in reducing the impact of IV, unless specific study designs are adopted that allow to capture multiple observations for each individuals across time~\cite{aarsand2018biological,bartlett2015checklist}. 

Despite these apparently relevant characteristics, the phenomenon of IV has largely been overlooked in the ML literature: indeed, while recent works have started to apply ML techniques to analyze IV data, for example to cluster patients based on their IV profiles~\cite{carobene2021multicenter} or to provide Bayesian models for IV~\cite{aarsand2021european}, to our knowledge no previous work has investigated the impact of IV on ML systems, as well as possible techniques to improve robustness and manage this source of perturbations.

In this article, we attempt to bridge this gap in the specialized literature, by addressing two main research problems. To this aim, this paper will consist of two parts: in the first part we will address the research question ``can individual variation significantly affect the accuracy, and hence the robustness, of a machine model on a diagnostic task grounding on laboratory medicine data'' ($H_{1}$). Due to the pervasiveness of individual variation, proving this hypothesis could suggest that most ML models could be seriously affected by lack of robustness on real-world and external data. To this aim, we will apply a biologically-grounded, generative model to simulate the effects of IV on data, and we will show how commonly used classes of ML models fail to be robust to it. On the other hand, the second part of the paper will aim to build on the rubble left by the first part, and it will address the hypothesis whether more advanced learning and regularization methods (grounding on, either, data augmentation~\cite{van2001art} or data imprecisiation~\cite{lienen2021instance}) will achieve increased robustness in face of the same perturbations ($H_{2}$).

\section{Background and Methods}
As discussed in the previous section, the aim of this article is to evaluate and address the potential impact of IV on ML models' robustness. 
In this section, we first provide basic background on IV, its importance in clinical settings, and methods to compute it. Then, in the next sections, we will describe two different experiments: in the first experiment, we evaluate how commonly used ML models fare when dealing with data affected by IV; then, in the second experiment, we evaluate the application of more advanced ML approaches to improve robustness to IV.

\subsection{Individual Variation in Medical Data}
\label{sec:background}
IV is considered one of the most important sources of uncertainty in clinical data~\cite{plebani2015biological} and recent research has highlighted the need to take IV properly into account in any use of medical data~\cite{badrick2021biological,frohlich2018hype}
. IV can be understood as encompassing three main components: pre-analytical variation, analytical variation and (within-subject) biological variation~\cite{fraser2001biological,plebani2015biological}.

Pre-analytical variation denotes uncertainty due to patients' preparation (e.g., fasting, physical activity, use of medicaments) or sample management (including, collection, transport, storage and treatment)~\cite{ellervik2015preanalytical}; it is usually understood that pre-analytic variation can be controlled by means of careful laboratory practice~\cite{fraser2001biological}. AV, by contrast, describes the un-eliminable uncertainty which is inherent to every measurement technique, and is characterized by both a random component (i.e., variance, that is the agreement between consecutive measurements taken with the same instrument); and a systematic component (i.e., bias, that is the differences in values reported by two different measurement instruments). Finally, BV describes the uncertainty arising from the fact that features or biomarkers can change through time, contributing to a variance in outcomes from the same individual that is independent of other forms of variation.

As already mentioned, IV can influence the interpretation and analysis of any clinical data: for this reason, quantifying IV, also in terms of its components, is of critical importance. However collecting reliable data about IV is not an easy task~\cite{carobene2018providing,haeckel2021problems}.
To this aim, standardized methodologies have recently been proposed~\cite{aarsand2018biological,bartlett2015checklist}: intuitively, IV can be estimated~\cite{aarsand2021european,carobene2018providing,roraas2012confidence} by means of controlled experimental studies that monitor \emph{reference individuals}\footnote{The term reference individual denotes an individual that, for some reasons, can be considered representative of the population of interest (e.g., healthy patients).}~\cite{carobene2016sample} by collecting multiple samples over time.

Formally speaking, let us assume that a given feature of interest $x$ has been monitored in $n$ patients for $m$ time steps. At each time step, $k \geq 2$ repeated measurements should be performed, so as to determine the AV component of IV. 
Then, the IV of feature $x$, for patient $i$, is estimated as $IV_i(x) = Variance(x^i)$, 
 while the AV component is defined as $AV_i(x) = Variance(x_s^i)$, 
 where $x^i$ denotes the collection of values of $x$ for patient $i$, and $x_s^i$ denotes the collection of values of $x$ for patient $i$ at the $s$-th time step. Then, the BV component of IV is computed as $BV_i(x) = \sqrt{IV_i(x)^2 - AV_i(x)^2}$.
Usually, IV, AV and BV are expressed in percent terms, defining the so-called coefficients of individual (resp., analytical, biological) variation, that is
  $CVT_i(x) = \frac{IV_i(x)}{\hat{x}^i}$,  $CVA_i(x) = \frac{AV_i(x)}{\hat{x}^i}$ and $CVI_i(x) = \frac{BV_i(x)}{\hat{x}^i}$.
 The overall variations, finally, can be computed as the average of the coefficients of variation across the population of patients.
The value of CVT, for a given set of features $x = (x_1, ..., x_d)$, can then be used to model the uncertainty about the observations obtained for any given patient $i$: indeed, any patient $i$, as a consequence of the uncertainty due to IV, can be represented by a d-dimensional Gaussian $\mathcal{N}_i(x^i, \Sigma^i)$, where $x^i$ is a $d$-dimensional vector characteristic representation of patient $i$, called \emph{value at the homeostatic point}, and $\Sigma^i$ is the diagonal covariance matrix given by $\Sigma^i_{j,j} = CVT(x_j)*x^i_j$~\cite{fraser2001biological}. More generally, having observed a realization $\hat{x}^i$ of $\mathcal{N}_i(x^i, \Sigma^i)$ for patient $i$, its distribution can be estimated as $\mathcal{N}_i(\hat{x}^i, \hat{\Sigma}^i)$, where $\hat{\Sigma}^i_{j,j} = CVT(x_j)*\hat{x}^i_j$.

Due to the complexity of design studies to obtain reliable IV estimates, a few compiled sources of IV data, for healthy patients, are available: the largest existing repositories in this sense, are the data originating from the European Biological Variation Study (EuBIVAS) and the Biological Variation Database (BVD)~\cite{aarsand2020eflm,sandberg2022biological}, both encompassing data about commonly used laboratory biomarkers. In the following sections, we will rely on data available from these sources in the definition of our experiments.

\subsection{Individual Variation and Statistical Learning}
\label{sec:rel}
One of the most simple yet remarkable results in Statistical Learning Theory (SLT) is the \emph{error decomposition theorem} \cite{shalev2014understanding} (also called bias-variance tradeoff, or bias-complexity tradeoff), which states that the true risk $L_D(h)$ of a function $h$ from a family $H$ w.r.t. to a distribution $D$ on the instance space $Z = X \times Y$ can be decomposed as:
\begin{equation}
\label{eq:decomp}
L_D(h) = \epsilon^{Bayes} + \epsilon^{Bias} + \epsilon^{Est}
\end{equation}

where $\epsilon^{Bayes} = min_{f \in F}L_D(f)$ is the \emph{Bayes error}, i.e. the minimum error achievable by any measurable function; $\epsilon^{Bias} = min_{h' \in H}L_D(h') - min_{f \in F} L_D(f)$ is the \emph{bias}, i.e. the gap between the Bayes error and the minimum error achievable in class $H$; $\epsilon^{Est} = L_D(h) - min_{h' \in H} L_D(h')$ is the \emph{estimation error}, i.e. the gap between the error achieved by $h$ and the minimum error achievable in $H$. 

A striking consequence of IV for ML tasks regards a generalization of the error decomposition theorem due to the impossibility of accessing the true distributional-valued representation of instances but only a sample drawn from the respective distributions. To formalize this notion, as in the previous section, denote with $f_i = \mathcal{N}(x^i, \Sigma^i)$ the distributional representation due to IV for instance $i$. Then, the learning task can be formalized through the definition of a \emph{random measure}~\cite{herlau2016completely} $\eta$ defined over the Borel $\sigma$-algebra $(Z, \mathcal{B})$ on the instance space $Z = X \times Y$, which associates to each instance $(x, y)$ a probability measure $\mathcal{N}(x, \Sigma) \times \delta_y$, where $\delta_y$ is the Dirac measure at $y \in Y$. A training set $S = \{ (x_1, y_i), \ldots, (x_m, y_m)\}$ is then obtained by first sampling $m$ random measures $f_1, \ldots, f_m$ from $\eta^m$, and then, for each $i$, by sampling a random element $(x_i, y_i) \sim f_i$. Then, the IV-induced generalization of the error decomposition theorem can be formulated as:
\begin{equation}
\label{eq:generaldecomp}
    L_\eta(h) = \epsilon^{Bayes}_\eta + \epsilon^{Bias}_\eta + \epsilon^{Est}_\eta + \epsilon^{IV}_\eta
\end{equation}

Indeed, the true error of $h$ w.r.t. $\eta$ can be expressed as $L_\eta(h) = E_{F \sim \eta^m}\left[\frac{1}{m}\sum_{f_i \in F} E_{(x_i, y_i) \sim f_i}l(h, (x_i, y_i))\right]$. Letting $D$  be the probability measure over $X \times Y$ obtained as the \emph{intensity measure} \cite{kallenberg2017random} of $\eta$, and $L_D(h) = E_{S \sim D^m} L_S(h)$ be the expected error of $h$ w.r.t. to the sampling of a training set $S$ from the product measure $D^m$, then the above expression can be derived by setting $\epsilon_\eta^{Bayes} = min_{f \in F} L_\eta(f)$, $\epsilon_\eta^{Bias} = min_{h' \in H} L_\eta(h') - min_{f \in F} L_\eta(f)$, $\epsilon_\eta^{Est} = L_D(h) - min_{h' \in H} L_\eta(h')$ and $\epsilon^{IV}_\eta = E_{F \sim \eta^m, S \sim D}\left[\frac{1}{m}\sum_i E_{(x_i, y_i) \sim f_i}l(h, (x_i, y_i)) - l(h, (x_i', y_i')\right]$.

Thus, compared with Eq \eqref{eq:decomp}, Eq \eqref{eq:generaldecomp} includes an additional error term $\epsilon^{IV}$ which measures the gap in performance due to the inability to use the IV-induced distributional representation of the instances, bur rather only a single instantiation of such distributions. This aspect is also reflected in the estimation error component in which the reference $min_{h' \in H} L_\eta(h')$ is compared not with the true error $L_\eta(h)$ but rather with the expected error over all possible instantiations $L_D(h)$. In the following sections, we will show, through an experimental study, that the impact of IV can be significant and lead to an overestimation of any ML algorithm's performance and robustness.

\subsection{Measuring the Impact of Individual Variation on Machine Learning Models}
\label{sec:dest}
In order to study whether and how the performance of a ML model could be impacted by IV, we designed an experiment through which we evaluated several commonly adopted ML models in the task of COVID-19 diagnosis from routine laboratory blood exams, using a public benchmark dataset. Aside from its practical relevance~\cite{cabitza2021development}, we selected this task for three additional reasons. First, blood exams are considered one of the most stable panels of exams~\cite{coskun2020systematic}: this allows us to evaluate the impact of IV in a conservative scenario where the features of interest are affected by relatively low levels of variability. Second, validated data about IV for healthy patients who underwent blood exams are available in the specialized literature~\cite{buoro2017short,buoro2017biological,buoro2018short} and these exams have high predictive power for the task of COVID-19 diagnosis~\cite{chen2021clinical}. Third, the selected dataset was associated with a companion longitudinal study~\cite{anonymousprognostic} that has been used to estimate IV data for the COVID-19 positive patients: we believe this to be particularly relevant since, even though IV data are available for healthy patients, no information of this kind is usually available for non-healthy patients, due to the complexity of designing studies for the collection of IV data, which could exhibit disease-specific patterns.
Although the estimation of IV is of paramount importance, both in medicine and other safety-critical domains, the striking lack of datasets presenting information to assess IV makes it a priority to devote further efforts and initiatives to make such resources available to the ML research community to make their models more robust and reliable. 

To this purpose, we used a dataset of patients who were admitted at the emergency department of the IRCCS Ospedale San Raffaele and underwent a COVID-19 test~\cite{anonymizedimportance}. The dataset was collected between February and May 2020 and encompasses 18 continuous features and 3 binary features (including the target). Since the dataset was affected by missing data, in order to limit the bias due to data imputation, we discarded all instances having more than 25\% missing values: the resulting dataset encompasses 1422 instances and is described in Table~\ref{tab:data} in Appendix A.

To evaluate the impact of IV, we used a biologically-informed generative model whose aim was to simulate the effect of biological and analytical variation on the measured features of the patients in the dataset. More in detail, based on the definition and computation of IV described previously, the generative model is defined by a case-dependent, class-conditional, multi-variate Gaussian distribution $N(x, \Sigma^{x,y})$, where we recall $\Sigma^{x,y} = diag(\langle x, \sqrt{CVA^2 + CVI_y^2} \rangle)$. We note that, even though the assumptions of normality and independence of variables may be considered strong, they are widely adopted in the specialized IV literature~\cite{fraser2001biological} as well as implicitly in the release format of the available IV data sources. Nonetheless, we believe that further work should be devoted at exploring more general models of IV that may take into account dependencies among features.

More in particular, for $CVA$ and $CVI_{y=0}$ we considered values previously reported in the literature~\cite{buoro2017biological,buoro2017short,buoro2018short}, while the values of $CVI_{y=1}$ were estimated from the longitudinal observation of the COVID-19 positive patients considered in this study~\cite{anonymousprognostic}, using the same methodology as described in the previous section. 

We considered 7 different ML models, commonly used in medical settings on tabular data, namely: Support Vector Machine (with RBF kernel) (SVM), Logistic Regression (LR), k-Nearest Neighbors (KNN), Naive Bayes (NB), Random Forest (RF), Gradient Boosting (GB), ExtraTrees (ET). We evaluated, in particular, the scikit-learn implementations of the previous models, with default hyper-parameters. Further information on implementation details is in Appendix C. We did not evaluate deep learning models as such models often require extensive hyper-parameter optimization and are usually out-performed by other models on tabular data \cite{grinsztajn2022tree}. The impact of IV on the performance of the above mentioned ML models was evaluated by means of a repeated cross-validation evaluation procedure: for a total of 100 iterations, a 3-fold cross-validation procedure was applied. More in detail, in each 3-fold cross-validation the two training folds were used to train the ML models, while the test fold $Te$ was used to obtain a perturbed fold $Te_p$ as follows: for each instance $(x,y) \in Te$, a perturbed instance $(x_p, y)$ was obtained to simulate the effect of individual variation, by sampling $x_p$ from $N(x, \Sigma^{x,y})$. The trained ML model was then evaluated on both $Te$ and $Te_p$ to measure the impact of individual variation, if any, by comparing the distribution of average performance on the original test folds with that of the perturbed test folds. 
In terms of performance metrics, we considered the accuracy, the AUC and the F1 score. The robustness of the ML models to IV was evaluated by comparing the average performance on the non-perturbed and IV perturbed data: in particular, we considered a model to be robust to IV if the 95\% confidence intervals for the above mentioned quantities overlapped (equivalently, the confidence interval of the difference included the value 0).

\subsubsection{Results}
\label{ssec:resdestr}
First of all, we assessed whether the IV perturbed data obtained by means of the considered generative model was statistically significantly different from the original data: ideally, to be realistic, IV-based perturbations of data should not influence too much the overall data distribution~\cite{fraser2009reference}.
To this purpose, we considered a subset of 4 predictive features (namely LY, WBC, NE and AST), which were previously shown to be among the most predictive features for the considered task~\cite{chen2021clinical}. We compared the distributions of the above mentioned features before and after the IV perturbations, by means of the Kolmogorov-Smirnov test with $\alpha = 0.01$. The obtained p-values were, respectively, 1 (for LY, WBC and NE) and 0.104 (for AST): thus, for all of the considered features, the null hypothesis of equal distributions for the IV perturbed and non-perturbed data could not be rejected.

\begin{figure*}[tbh]
  \centering
  \includegraphics[width=0.75\textwidth]{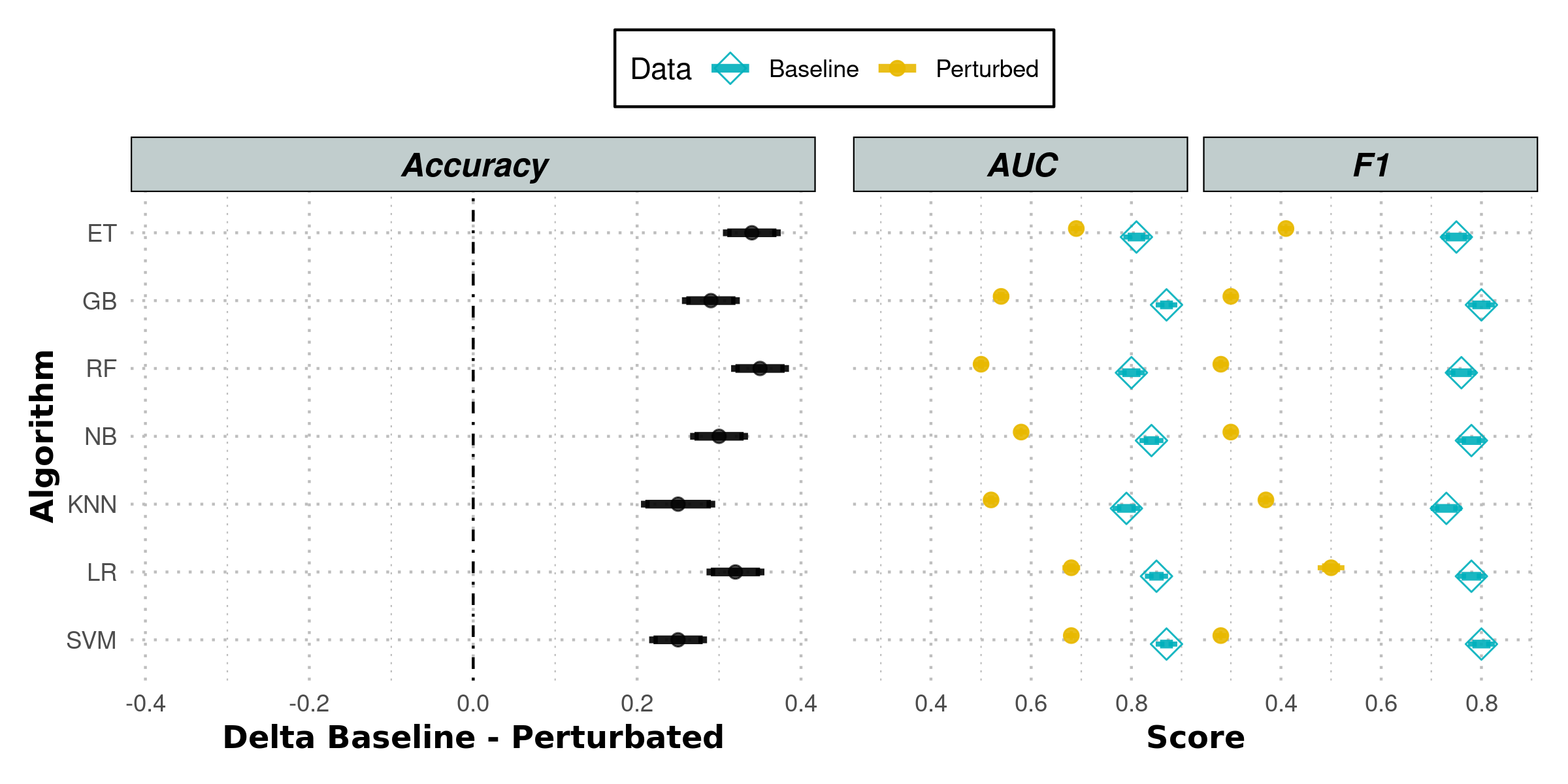}
  \caption{Results of the experiments for measuring the impact of IV on the performance of standard ML models. For each algorithm and metrics, we report the average and 95\% confidence interval for both baseline (that is, non-perturbed) and IV perturbed data.}
  \label{fig:destruens}
\end{figure*}

The impact of IV on the ML models is reported in Figure~\ref{fig:destruens}.
The difference in performance (baseline vs perturbed) was significant for all algorithms: indeed, for all algorithms, the confidence intervals on the baseline and IV perturbed data did not overlap. The best algorithms on the non-perturbed data were RF and ET, w.r.t. all considered metrics (AUC: 0.87, Accuracy: 0.8, F1: 0.8); while the best algorithms on the the IV perturbed data were SVM (w.r.t. AUC: 0.69, and Accuracy: 0.5) and GB (w.r.t. F1: 0.5).

These results highlight how, even though the distributions of highly predictive feature were not significantly affected by IV, IV nonetheless had a significant impact on the performance of the considered ML algorithms, that were therefore not robust to IV-related uncertainty. Algorithms, however, were not equal in their robustness (or lack thereof) w.r.t. IV: in particular, the more robust models were SVM (w.r.t. Accuracy, with average performance decrease 0.25, and AUC, with average decrease 0.12) and GB (w.r.t. F1 score, with average performance decrease 0.28), with all other models being significantly less robust (that is, having a significantly larger difference between baseline and IV perturbed performances). While this latter observation can be given a learning theoretical justification based on the notion of margin\footnote{Both SVM and GB are margin-based classifiers~\cite{gronlund2020margins,hanneke2021stable}. It is not hard to see that the existence of a large margin on the non-perturbed data is a necessary (but not sufficient) condition for robustness to IV.}, we note that even SVM and GB reported a significant decrease in performance on the IV-perturbed data: thus, even models that are usually considered to be robust to noise can nevertheless be strongly affected by IV.

\subsection{Data Augmentation and Imprecisiation Methods to Manage Individual Variation}
\label{sec:constr}
In light of the results reported in the previous section, which show the lack of robustness of standard ML models w.r.t. IV, in this section we investigate the application of more advanced methods that attempt to directly address the representation of IV in data and hence tackle the error decomposition show in Eq \eqref{eq:generaldecomp}. In particular, we consider approaches based either on \emph{data augmentation} or \emph{data imprecisiation}. In both cases, we adopted the same experimental protocol described in the previous section (see below).

Data augmentation~\cite{chen2021synthetic,van2001art} refers to regularization techniques that aim to increase the stability and robustness of a ML model by enriching the training set with new instances. In our setting, the idea is to inject further information related to the IV distribution within the model to improve generalization. 

Since in the considered setting a generative model of IV was available, this latter was used to generate synthetic data points to augment the original training set. Basically, for each instance $(x,y)$ in the training folds, we generated $n=100$ new samples from the distribution $N(x, \Sigma^{x,y})$, so as to simulate the effect of having multiple observations, perturbed by IV, for each patient. We considered, in particular, the application of the above mentioned basic data augmentation strategy to the SVM (denoted as ACS) and Gradient Boosting (denoted as ACG) ML models, since these latter two were shown to be more robust to IV (see previous section). The pseudo-code for evaluating the data augmentation models is reported in Algorithm~\ref{algo:augm}.

\begin{algorithm}[!htb]
\begin{algorithmic}
\Procedure{data\_augmentation\_eval}{$h$: ML model, $S$: dataset, $M$: metric, $n$ : number of augmented instances}
\ForAll{iterations $i=1$ to $100$}
  \State Split $S$ in 3 class-stratified folds
  \ForAll{$Tr$: training fold, $Te:$ test fold}
    \State $Tr_a = \emptyset$
    \ForAll{$(x, y) \in Tr$}
      \ForAll{iteration $j=1$ to $n$}
        \State Add to $Tr_a$  $(x_p, y)$, $x_p \sim N(x, \Sigma^{x,y})$
      \EndFor
    \EndFor
    \State $Te_p = \emptyset$
    \ForAll{$(x, y) \in Te$}
      \State Add to $Te_p$  $(x_p, y)$, $x_p \sim N(x, \Sigma^{x,y})$
    \EndFor
    \State Train $h$ on $Tr_a$
    \State Eval $h$ on $Te$ ($M(h, Te)$), $Te_p$ ($M(h, Te_p)$)
  \EndFor
\EndFor
\State \textbf{return} The distributions of $M(h, Te)$ and $M(h, Te_p)$
\EndProcedure
\end{algorithmic}
\caption{The procedure to evaluate the impact of IV on the data augmentation-based ML models.}
\label{algo:augm}
\end{algorithm}

By contrast, data imprecisiation~\cite{hullermeier2014learning,lienen2021instance} refers to ML techniques by which data affected by some form of uncertainty are transformed into imprecise observations, that is distributions over possible instances, which are then used to train specialized ML algorithms. Formally speaking, an \emph{imprecisiation scheme} is a function $i: X \times Y \mapsto [0,1]^{X\times Y}$, where $X$ is the feature space. In the experiments, we considered two commonly adopted imprecisiation schemes grounding on, respectively, probability theory and possibility theory~\cite{denoeux2020representations}, namely:
\begin{align}
  i_{prob} &: (x, y) \mapsto (N(x, \Sigma^{x,y}), y)\\
  i_{poss} &: (x, y) \mapsto (Gauss(x, \Sigma^{x,y}), y) 
\end{align}
where $Gauss(a, b)$ denotes the Gaussian fuzzy vector, whose $j$-component is defined as $Gauss(a,b)_j(x) = e^{\frac{(x-a)^2}{b^2}}$. Intuitively, $i_{prob}$ represents each instance affected by IV as a Gaussian probability distribution over possible instances, while $i_{poss}$ represents each instance affected by IV as a Gaussian possibility distribution (equivalently, a Gaussian fuzzy set) over possible instances. Thus, the general idea of applying data imprecisiation (and corresponding ML algorithms) in our setting is to model the uncertainty due to IV by representing each instance as a cloud of points in the feature space whose distribution is determined by the IV parameters, as a form of regularization.

We considered three ML algorithms proposed in the learning from imprecise data literature, namely: k-Nearest Distributions (KND, also called Generalized kNN)~\cite{zheng2010k}, Support Measure Machine (SMM)~\cite{muandet2017kernel}, Weighted re-Sampling Forest (WSF)~\cite{seveso2020ordinal}. See also Appendix C for hyper-parameter settings for the considered models.
KND denotes the generalization of kNN to distribution-valued instances, namely we used the $i_{prob}$ scheme\footnote{Since Mahalanobis' distance takes into account only the mean and scale, using $i_{poss}$ scheme would result in the same algorithm.} and Mahalanobis distance:
\begin{equation}
\begin{split}
(x_1 - x_2)^T \frac{{\Sigma^{x_1, y_1}}^{-1} + {\Sigma^{x_2, y_2}}^{-1}}{2}(x_1 - x_2)
\end{split}
\end{equation}

SMM, by contrast, refers to the generalization of SVM to instances represented as probability distributions (thus, only the $i_{prob}$ imprecisiation scheme was considered). The SMM model grounds on the notion of a \emph{kernel mean embedding}~\cite{muandet2017kernel}, that is a generalization of the notion of kernel in ML to the the space of probability distributions, which could thus be seen as a measure of similarity between two imprecise instances. For computational complexity reasons, we considered the RBF kernel, which for normally distributed imprecise instances can be expressed in closed form as~\cite{muandet2017kernel}:
\begin{equation}
\label{eq:rbf}
\begin{split}
RBF_{\gamma, i_{prob}} = \frac{e^{-\frac{(x_1 - x_2)^T (\Sigma^{x_1, y_1} + \Sigma^{x_2, y_2} + \frac{1}{\gamma}I)^{-1}(x_1 - x_2)}{2}}}{\sqrt{det(\gamma\Sigma^{x_1, y_1} + \gamma\Sigma^{x_2, y_2} + I)}}
\end{split}
\end{equation}

Finally, the WSF model is an approximation algorithm to solve the generalized risk minimization problem~\cite{hullermeier2014learning}, a commonly adopted~\cite{lienen2021label,lienen2021instance} approach to deal with imprecise. WSF is based on a generalization of bootstrapped tree ensembles to instances represented as possibility distributions (thus, only the $i_{poss}$ imprecisiation scheme was considered): in addition to the randomization w.r.t. the split point selection and the bootstrap re-sampling of the instances, an additional randomization on the feature values is considered. Specifically, for each tree in the ensemble, each imprecise instance $i_{poss}(x, y)$ in the corresponding bootstrap set is used to sample an instance $(x', y')$, by means of a two-step procedure~\cite{dubois1993possibility}: first, a number $\alpha \in [0,1]$ is selected uniformly at random, then a random value is drawn from the $\alpha$-cut $i_{poss}(x, y)^\alpha = \{ (x', y') \in X \times Y : i_{poss}(x, y)(x', y') \geq \alpha \}$. A pseudo-code description of WSF, along with an analysis of its computational complexity and generalization error, is in Appendix B.

The KND, SMM and WSF models were implemented in Python, and evaluated in a setup similar to the one adopted for the data augmentation-based ML models, as shown in Algorithm~\ref{algo:imprec}. The full code for the algorithms and evaluation procedures is available on GitHub at \url{anonymized url}.

\begin{algorithm}[!htb]
\begin{algorithmic}
\Procedure{data\_imprecisiation\_eval}{$h$: ML model, $S$: dataset, $M$: metric, $i$: imprecisiation scheme}
\ForAll{iterations $t=1$ to $100$}
  \State Split $S$ in 3 class-stratified folds
  \ForAll{$Tr$: training fold, $Te:$ test fold}
    \State $Tr_a = \emptyset$; $Te_b = \emptyset$; $Te_p = \emptyset$
    \ForAll{$(x, y) \in Tr$}
      \State $Tr_a.append(i((x,y)))$
    \EndFor
    \ForAll{$(x, y) \in Te$}
      \State $Te_b.append(i((x, y)))$
      \State Sample $(x_p, y) \sim N(x, \Sigma^{x,y})$
      \State $Te_p.append(i((x_p, y)))$
    \EndFor
    \State Train $h$ on $Tr_a$
    \State Eval $h$ on $Te_b$ ($M(h, Te_b)$), $Te_p$ ($M(h, Te_p)$)
  \EndFor
\EndFor
\State \textbf{return} The distributions of $M(h, Te_b)$ and $M(h, Te_p)$
\EndProcedure
\end{algorithmic}
\caption{The procedure to evaluate the impact of IV on the data imprecisiation-based ML models.}
\label{algo:imprec}
\end{algorithm}

\subsubsection{Results}
\label{ssec:resconstr}
\begin{figure*}[tbh]
  \centering
  \includegraphics[width=0.75\textwidth]{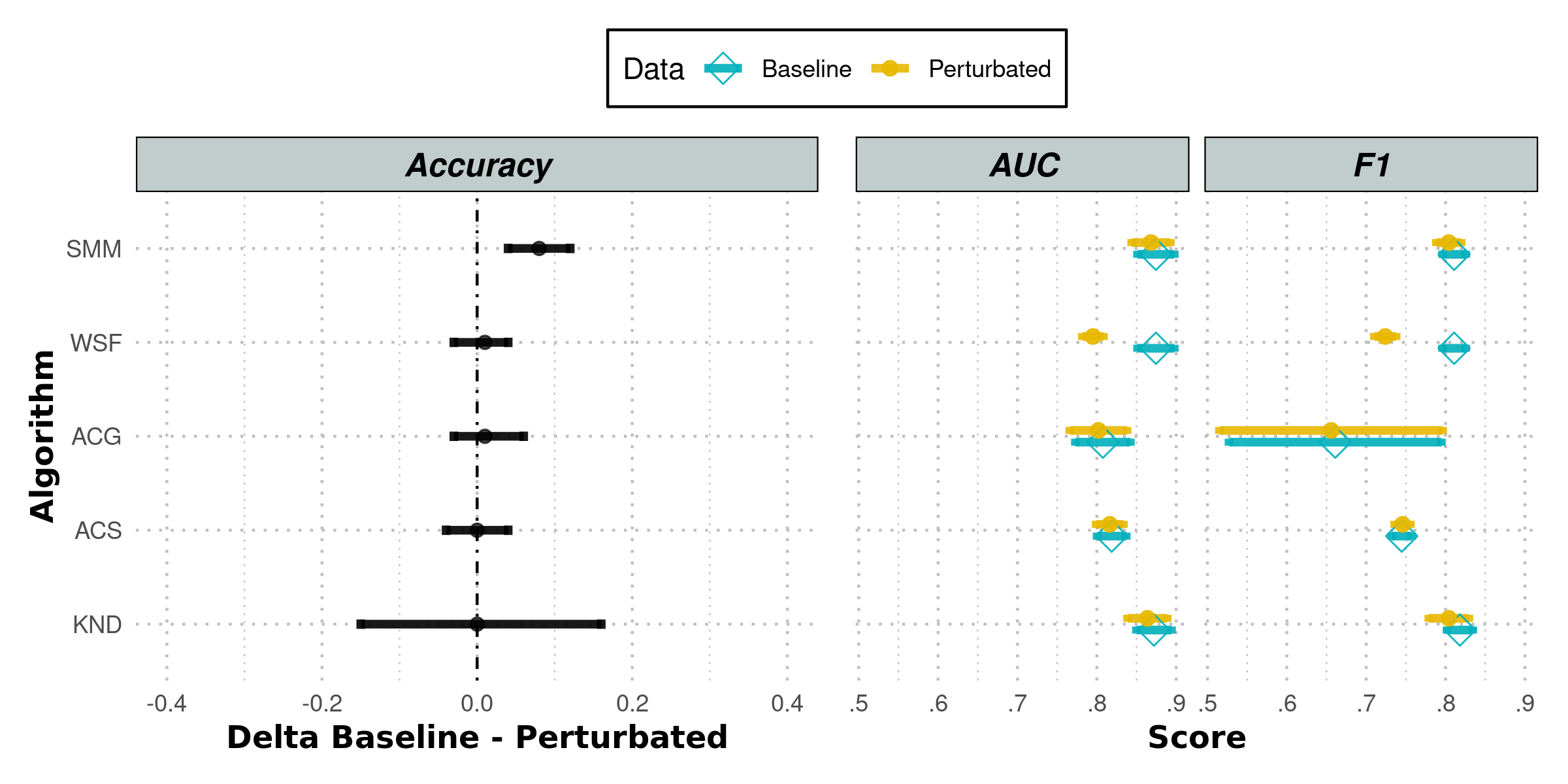}
  \caption{Results of the experiments for measuring the impact of IV on the performance of data augmentation-based and data imprecisiation-based ML models. For each algorithm and metrics, we report the average and 95\% confidence interval for both baseline (that is, non-perturbed) and IV perturbed data.}
  \label{fig:construens}
\end{figure*}

The results for data augmentation and imprecisiation-based ML models are reported in Figure~\ref{fig:construens}.
For all models except SMM, the difference in performance on baseline and IV perturbed data was not significant. The best models on the non-perturbed data were SMM, WSF (w.r.t. AUC: 0.87) and WSF, ACG (w.r.t. Accuracy: 0.8, F1: 0.81), while the best models on the IV perturbed data were ACG and WSF (AUC: 0.86, Accuracy: 0.79, F1: 0.8). Comparing these results with those shown in the previous section, it is easy to observe that both data augmentation and data imprecisiation-based ML models were much more robust to IV perturbations than the standard ML models. Indeed, the most robust models (w.r.t. AUC: WSF and ACS, with average difference 0.003; w.r.t. Accuracy and F1: WSF and ACG, with average difference 0.006) were hardly impacted by IV. Even the least robust model (i.e., SMM) was much more robust than the standard ML models (average differences w.r.t. AUC: 0.08; w.r.t. Accuracy: 0.09, w.r.t. F1: 0.09).

In light of these results, we claim that data augmentation and imprecisiation can be helpful to improve robustness under IV perturbations. We conjecture this to be due to directly taking into account information about IV in data representation and model training, which allows to strike a trade-off among the various components of the generalized error decomposition shown in Eq. \eqref{eq:generaldecomp}. We note that these two approaches, while performing similarly in terms of accuracy and robustness, have different characteristics that may influence their suitability in practical scenarios. Data augmentation methods allow to use out-of-the-box ML models, since IV management is implemented as a pre-processing step: this is not the case for data imprecisiation-based approaches, which require specialized ML algorithms. By contrast, imprecisiation-based approaches have lower computational complexity and may thus scale better on larger datasets: e.g., if $m$ is the training set size, $d$ the number of features, $n$ the number of ensembled models, and $r$ the number of augmented instances then, the time complexities of SMM and WSF are, respectively, $O(m^2d^3)$ and $O(n\,d\,m\log(m))$; by contrast, the complexities of ACS and ACG are, respectively, $O(m^2r^2)$ and $O(n\,d\,m\,r\log(m r))$.

\section{Conclusion}
\label{sec:concl}
In this article we studied the impact of IV, an oft neglected type of uncertainty affecting data, on the performance and robustness of ML models. Crucially, through a realistic experiment on COVID-19 diagnosis, we showed that standard ML algorithms can be strongly impacted by the presence of IV, failing to generalize properly. Such an issue can severely limit the applicability and safety of ML methods in tasks where data are expected to be affected by IV, that is most applications in clinical settings and more generally in real-world domains where the manifestations of the phenomena of interest could exhibit time-varying patterns. Our results then imply that out-of-the-box methods cannot be naively applied in such domains. Nonetheless, every cloud has a silver lining, and we showed that more advanced learning methods, grounding on data augmentation and data imprecisiation, can achieve better robustness w.r.t. IV: this highlights the need to employ models that take into account the \emph{generative history} underlying the data acquisition process, including the uncertainty due to IV, in their learning algorithms. Furthermore, we believe that our results highlight the importance of adopting proper algorithmic and experimental designs for ML studies in medicine: due to the potential impact of IV on the performance of ML models, data collection studies should be designed so as to enable the estimation of IV values which could then be used in the ML development phase. Thus, increasing emphasis should be placed on longitudinal studies, or otherwise studies in which multiple samples are collected for each involved patients under controlled conditions, so as to allow precise and reliable estimation of IV. We believe that these results could pave the way for the investigation of IV and its effects on the safety of ML models deployed in real-world clinical settings. Thus, we think that the following open problems could be of interest:
\begin{itemize}
  \item In our experiments we assumed the IV distributions to be Gaussian with diagonal covariance. While this model is commonly adopted in the literature, we believe that further research should explore the relaxation of this assumption, by considering more general models of IV accounting for causal relationships among features;
  \item While we focused on the impact of IV in ML and briefly discussed IV in SLT, we believe the theoretical side of this issue merits further study: even though the problem of learning from distributional data has recently been investigated in SLT \cite{campagner2021learnability,ma2021learning,muandet2017kernel}, this area is still in its infancy;
  \item In this work we showed the impact of IV in the setting of COVID-19 diagnosis from blood tests. Future work should generalize our analysis to a broader spectrum of applications.
\end{itemize}

\clearpage
\appendix
\setcounter{table}{0}
\renewcommand{\thetable}{A\arabic{table}}
\setcounter{figure}{0}
\renewcommand{\thefigure}{A\arabic{figure}}
\setcounter{algorithm}{0}
\renewcommand{\thealgorithm}{A\arabic{algorithm}}

\section{Appendix A: Data Characteristics}
Descriptive statistics for the considered dataset are reported in Table \ref{tab:data}.

\begin{table}[htb]
 \centering
 \caption{The list of features, along with the target. Mean and standard deviation are reported for continuous features, distribution of values is reported for discrete feature. For the discrete features we report the distribution of values. For the laboratory blood data, we also report the analytical (CVA) and biological (CVI) variation, differentiated by healthy vs non-healthy patients, and missing rate.}
 \label{tab:data}
 \setlength{\tabcolsep}{0.1em}
 \resizebox{\columnwidth}{!}{
\begin{tabular}{c|c|c|c|c|c|c|c|c}
\hline
         Features & Acronym &           Units &  Mean &  Std & Missing & CVA & CVI$_{y=0}$ & CVI$_{y=1}$ \\\hline
\hline
  \makecell{Alanine\\Transaminase} &   ALT &            U/L & 39.87 & 42.26 &   0.07 & 0.04 & 0.093 & 0.051 \\\hline
  \makecell{Aspartate\\Transaminase} &   AST &            U/L & 46.90 & 51.90 &   0.14 & 0.04 & 0.095 & 0.52\\\hline
  \makecell{Alkaline\\Phosphatase} &   ALP &            U/L & 88.61 & 72.09 &  16.24 & 0.05 & 0.054 & 0.045\\\hline
\makecell{Gamma\\Glutamyl\\Transferase} &   GGT &            U/L & 67.48 & 140.52 &  17.09 & 0.035 & 0.089 & 0.036\\\hline
  \makecell{Lactate\\Dehydrogenase} &   LDH &            U/L & 332.52 & 218.43 &   8.02 & 0.03 & 0.052 & 0.024\\\hline
      \makecell{Creatine\\Kinase} &   CK &            U/L & 184.47 & 382.02 &  56.19 & 0.05 & 0.145 & 0.062\\\hline
          Calcium &   CA &           mg/dL &  2.20 &  0.17 &   0.84 & 0.03 & 0.018 & 0.018\\\hline
         Glucosium &   GLU &           mg/dL & 119.12 & 55.80 &   0.42 & 0.028 & 0.047 & 0.026\\\hline
          Urea &  UREA &           mg/dL & 48.64 & 42.69 &  31.01 & 0.03 & 0.141 & 0.035\\\hline
        Creatinine &  CREA &           mg/dL &  1.19 &  1.01 &   0.07 & 0.025 & 0.044 & 0.022\\\hline
     \makecell{Leukocytes} &   WBC &           10$^9$/L &  8.65 &  4.77 &   0.00 & 0.019 & 0.111 & 0.033\\\hline
      \makecell{Erythrocytes} &   RBC &          10$^{12}$/L &  4.55 &  0.72 &   0.00 & 0.009 & 0.018 & 0.010\\\hline
        Hematocrit &   HCT &             \% & 39.47 &  5.57 &   0.00 & 0.018 & 0.024 & 0.019\\\hline
        Neutrophils &   NE &             \% & 72.48 & 13.35 &   8.51 & 0.03 & 0.146 & 0.014\\\hline
        Lymphocytes &   LY &             \% & 18.58 & 11.11 &   8.51 & 0.036 & 0.11 & 0.043\\\hline
         Monocytes &   MO &             \% &  7.76 &  3.86 &   8.51 & 0.063 & 0.134 & 0.033\\\hline
        Eosinophils &   EO &             \% &  0.82 &  1.59 &   8.51 & 0.079 & 0.156 & 0.098\\\hline
         Basophils &   BA &             \% &  0.34 &  0.27 &   8.51 & 0.031 & 0.128 & 0.056\\\hline
            Sex &   - &            \makecell{Female\\Male} &  \makecell{42\%\\58\%} &  - &   - & - & - & -\\\hline
            Age &   - &           Years & 61.19 & 18.89 &   - & - & - & -\\\hline
          Target & - & \makecell{Positive\\Negative} &  \makecell{53\%\\47\%} &  - &   - & - & - & -\\\hline
\hline
\end{tabular}
}
\end{table}

Complete Blood Count data (i.e. features WBC, RBC, HCT, NE, LY, MO, EO, BA) was obtained by analysis of whole blood samples by means of a Sysmex XE 2100 haematology automated analyser. Biochemical data (ALT, AST, ALP, GGT, LDH, CK, CA, GLU, UREA, CREA) was obtained by analysis of serum samples by means of a Cobas 6000 Roche automated analyser. For each of the considered patients, COVID-19 positivity was determined based on the result of the molecular test for SARS-CoV-2 performed by RT-PCR on nasopharyngeal swabs: on a set of 165 cases for which the RT-PCR reported uncertain results, chest radiography and X-rays were also used to improve over the sensitivity of the RT-PCR test by combination testing.

\section{Appendix B: The WSF Algorithm}
Pseudo-code for the WSF algorithm is reported in Algorithm \ref{algo:wsf}. As described in the main text, the computational complexity of WSF is $O(n\,d\,|S|\log(|S|))$ where $d$ is the dimensionality of the input space.

\begin{algorithm}[thb]
\begin{algorithmic}
\Procedure{WSF}{$S$: dataset, $n$: ensemble size, $\mathcal{H}$ model class}
\State $Ensemble \gets \emptyset$
\ForAll{iterations $i=1$ to $n$}
  \State Draw a boostrap sample $S'$ from $S$
  \State $Tr_i \gets \emptyset$
  \ForAll{$(x,y) \in S'$}
    \State Sample $\alpha \sim U[0,1]$
    \State Add $(x',y') \sim i_{poss}(x, y)^\alpha$ to $Tr_i$
  \EndFor
  \State Add base model $h_i \in \mathcal{H}$ trained on $Tr_i$ to $Ensemble$
\EndFor
\State \textbf{return} $Ensemble$
\EndProcedure
\end{algorithmic}
\caption{The WSF algorithm.}
\label{algo:wsf}
\end{algorithm}

In regard to the generalization error of WSF w.r.t. data generating random measure, for each base model $h_i$, let $$L_{S}(h_i) = \sum_{(x,y) \in S} \mathbb{E}_{(x',y) \sim i_{poss}(x,y)}\left[\mathbbm{1}_{h(x') \neq y}\right]$$ and $L_D(h_i) = \mathbb{E}_{S \sim D^m} L_S(h)$, where $D$ is the intensity measure describe in Section ``\nameref{sec:rel}''. Assume further, that for all $h \in \mathcal{H}$, with probability larger than $1 - \delta$ if $(x - x')\Sigma^{x}(x-x') \leq T_{d, |S| - d}^2(1 - \delta)$ it holds that $h(x) = h(x')$, where $T$ is Hotelling's T-squared distribution \cite{hotelling1992generalization}. Intuitively, this latter condition can be understood as a strong form of regularity for models in $\mathcal{H}$: if two instantiations likely come from the same distribution due to IV, then with high probability they will be classified in the same way by each $h \in \mathcal{H}$.
Then, letting $V_i$ be the out-of-bag sample for model $h_i$, by Hoeffding's inequality and above assumptions it follows that, with probability $1 - \delta$, $L_D(h_i) \leq L_{V_i}(h_i) + \sqrt{\frac{\log(2|V_i|/\delta)}{2|V_i|}}$. Let $p = \sum_i L_{V_i}(h_i) + \sqrt{\frac{\log(2|V_i|/\delta)}{2|V_i|}} \leq \frac{1}{2}$. Then, assuming the $h_i$ err independently of each other, and noting that $WSF$ errs on an instance $x$ iff at least n/2 base models err, with probability greater than $1 - \frac{\delta}{n}$ the generalization error of WSF can be upper bounded through an application of Chernoff's bound for binomial distributions \cite{arratia1989tutorial} by $e^{-n \cdot KL(\frac{1}{2} || p)}$, where $KL(a || b) = a \log \frac{a}{b} + (1-a) \log \frac{1-a}{1-b}$ is the Kullback-Leibler divergence.

\section{Appendix C: Implementation Details and Hyper-parameter Settings}
All code was implemented in Python v. 3.10.4, using numpy v. 1.23.0, scikit-learn v. 1.1.1 and scikit-weak v. 0.2.0. For the standard ML models, we considered the default hyper-parameter values as defined in scikit-learn v. 1.1.1, with the exception of the random\_state seed, which was set to 99 for all evaluated models to ensure reproducibility, and the max\_depth hyperparameters for Random Forest and Gradient Boosting, which were set to 10 to avoid over-fitting and reduce the running time. For the data augmentation models we set the number of augmentation rounds to 100: for ACS we used as base model a SVC with rbf kernel and default hyper-parameters, while for ACG we used a GradientBoostingClassifier with max\_depth set to 0 and random\_state set to 99 for consistency with the classical case. For SMM we used as kernel the RBF kernel defined in \eqref{eq:rbf} with $\gamma = \frac{1}{\text{num. features}}$, while for WSF we used ExtraTreeClassifier as base classifier, we set the number of ensembled models to 100 and the random\_state seed to 99. Finally, for KND we set the number of neighbors $k$ to 5.

\end{document}